\documentclass[]{style/ceurart}
\usepackage{subcaption}
\usepackage{listings}
\begin{document}
\copyrightyear{2025}
\copyrightclause{Copyright for this paper by its authors.
  Use permitted under Creative Commons License Attribution 4.0
  International (CC BY 4.0).}

\conference{CLEF 2025 Working Notes, 9 -- 12 September 2025, Madrid, Spain}

\title{DS@GT AnimalCLEF: Triplet Learning over ViT Manifolds with Nearest Neighbor Classification for Animal Re-identification}
\title[mode=sub]{Notebook for the LifeCLEF Lab at CLEF 2025}

\author[1]{Anthony Miyaguchi}[
orcid=0000-0002-9165-8718,
email=acmiyaguchi@gatech.edu,
]
\cormark[1]

\author[1]{Chandrasekaran Maruthaiyannan}[
orcid=0009-0005-5757-4480,
email=chand2020@gatech.edu
]

\author[1, 2]{Charles R. Clark}[
orcid=0009-0008-0415-111X,
email=cclark339@gatech.edu
]
\cormark[1]

\address[1]{Georgia Institute of Technology, North Ave NW, Atlanta, GA 30332}
\address[2]{University of Florida, Gainesville, FL 32610}
\cortext[1]{Corresponding author.}

\begin{abstract}
This paper details the DS@GT team's entry for the AnimalCLEF 2025 re-identification challenge. 
Our key finding is that the effectiveness of post-hoc metric learning is highly contingent on the initial quality and domain-specificity of the backbone embeddings.
We compare a general-purpose model (DINOv2) with a domain-specific model (MegaDescriptor) as a backbone.
A K-Nearest Neighbor classifier with robust thresholding then identifies known individuals or flags new ones. 
While a triplet-learning projection head improved the performance of the specialized MegaDescriptor model by 0.13 points, it yielded minimal gains (0.03) for the general-purpose DINOv2 on averaged BAKS and BAUS.
We demonstrate that the general-purpose manifold is more difficult to reshape for fine-grained tasks, as evidenced by stagnant validation loss and qualitative visualizations. 
This work highlights the critical limitations of refining general-purpose features for specialized, limited-data re-ID tasks and underscores the importance of domain-specific pre-training.
The implementation for this work is publicly available at \url{github.com/dsgt-arc/animalclef-2025}.
\end{abstract}

\begin{keywords}
    Animal Re-identification \sep
    Open-Set Re-identification \sep
    Triplet Learning \sep
    Metric Learning \sep
    Vision Transformer (ViT) \sep
    DINOv2 \sep
    MegaDescriptor \sep
    Nearest Neighbor Classification \sep
    Kaggle \sep
    LifeCLEF \sep
    DS@GT
\end{keywords}

\maketitle
\section{Introduction}

Individual animal identification is helpful for biologists studying animal populations in the wild.
The ability to track an animal over time in its natural environment gives insights into behaviors and ecological interactions that are not possible with general census statistics.

In this paper, we describe the solution developed by the DS@GT team for the AnimalCLEF 2025 challenge hosted on Kaggle.
We utilize pre-trained, self-supervised vision transformers to embed animal images into embedding space and run a K-NN classifier with statistical thresholding to determine a label for each image.
We further refine the manifold learned by the vision transformer using a triplet learning procedure that learns to map individuals in space more effectively, achieved by projecting triplets of images from the ViT embedding space to a new projection for the metric. 
We hypothesize that a domain-specific backbone (MegaDescriptor) will provide a more suitable initial embedding manifold for triplet-based refinement than a general-purpose backbone (DINOv2), leading to greater performance gains on this specialized re-ID task.
Our method can overcome a simple baseline provided by the competition organizers, but further work is necessary.
\section{Related Work}

\subsection{Animal Re-identification} 

Animal re-identification (re-ID) refers to a system's ability to predict the identity of an individual animal based on its unique physical traits \cite{animalclef2025}. It is critical for biologists and ecologists to monitor populations, track movement, and study social behavior \cite{vcermak2024wildlifedatasets}. Approaches differ in their descriptor strategies. Early methods rely on local descriptors, such as SIFT, SURF, or contour features extracted from key points, to identify individuals via match counts \cite{vcermak2024wildlifedatasets}. More recent approaches use deep neural networks with metric learning to generate feature embeddings for identity matching \cite{vcermak2024wildlifedatasets}. Hybrid pipelines combine object detection and feature extraction to localize animals (or faces) before identifying them \cite{laskowski2023gorillavision, miele2021}. Numerous datasets support benchmarking: ATRW includes 3,649 images of 92 Amur Tigers in zoos \cite{atrw2019}; zebrafishRe-ID offers 2,224 images of 6 zebrafish in lab settings \cite{haurum_re-identification_2020}; and Cows2021 provides 13,784 images of 182 cows on a farm \cite{gao2021selfsupervisionvideoidentificationindividual}. The WildlifeReID-10K dataset aggregates 36 wildlife re-ID datasets, comprising approximately 140,000 images from over 10,000 individuals across multiple species \cite{wildlifereid10k}. Although many studies frame re-ID as a closed-set task, this assumption often breaks down in ecological settings where new individuals may appear.

\subsection{Vision Transformers for Computer Vision}

Dosovitskiy \textit{et al.} introduced the Vision Transformer (ViT), adapting the Transformer architecture to image patches, which achieved competitive performance with less computing when pre-trained on large datasets \cite{dosovitskiy2021an, vaswani2017}. However, quadratic scaling in image size led Liu \textit{et al.} to propose the Swin Transformer, which utilizes shifted-window attention for hierarchical, linear-complexity feature extraction and achieves strong performance across dense vision tasks \cite{swin2021liu}. In re-ID, vision transformers have been adopted to capture both short- and long-range features. TransReID was the first ViT-based method for person re-ID \cite{he2021transreid}, while GorillaVision applied a pre-trained ViT backbone to gorilla face recognition \cite{laskowski2023gorillavision}.

More recently, self-supervised and multi-modal transformer models have become powerful tools for vision tasks. DINOv2, a self-supervised ViT trained via self-distillation, performs well on fine-grained tasks such as species recognition \cite{oquab2023dinov2, miyaguchi2024transferlearningselfsupervisedvision}. CLIP, trained on image-text pairs, learns image embeddings that generalize well across domains, including re-ID \cite{radford2021clip, wu2024individualidentitydrivenframeworkanimal}. MegaDescriptor, a Swin-based model trained on a large multi-species re-ID dataset, achieves state-of-the-art performance across animal re-ID benchmarks, outperforming models like DINOv2 and CLIP \cite{vcermak2024wildlifedatasets}.

\subsection{Metric Learning}

Metric learning underpins many re-ID approaches. Triplet loss trains models on anchor-positive-negative triplets to ensure embeddings of same-identity pairs are closer than those of different identities \cite{facenet}. Popularized initially in face recognition, triplet learning is widely used in re-ID. To enhance separability, angular and margin-based losses such as ArcFace introduce additive angular margins on the hypersphere \cite{deng2019arcface}, with adaptations for re-ID. Recent methods, such as Matryoshka Representation Learning (MRL), produce hierarchical embeddings that encode coarse-to-fine details, enabling the model to dynamically select embedding subspaces for efficient nearest-neighbor search, depending on the retrieval task.

\section{Methodology}

Our experimentation is structured into two major phases, each serving a distinct purpose in our research.
Our first experiment validates the relative performance between DINOv2 \cite{oquab2023dinov2} and MegaDescriptor \cite{vcermak2024wildlifedatasets} using a K-NN classifier.
We use the pre-trained models in a zero-shot fashion and tune the threshold distance for new identities in the model using our dataset split.
In our second set of experiments, we undertake the task of reshaping the manifold. This involves a deliberate effort to bring images of the same animal closer together and push images of different animals further apart.
The goal is to disambiguate individuals and to make thresholds more robust.

\subsection{Competition Evaluation Metric}

The competition we participate in employs two specialized evaluation metrics, each of which plays a crucial role in assessing the performance of our models.
The first is Balanced Accuracy on Known Samples (BAKS) which measures the ability to identify individuals present in the training set.
The second is the Balanced Accuracy on Unknown Samples (BAUS) which is the ability to classify new individuals not in the training set as unknown.

More formally, we define the set of individuals $I$ into known subset $I_K$ and unknown subset $I_U$.
$N_c$ is the the number of images for an individual $c$ in set $I$.

\begin{equation} \label{eq:baks}
    \text{BAKS}(y, \hat{y}) = \frac{1}{|I_K|} \sum_{c \in I_K}
    \left(
        \frac{1}{n_c} \sum_{i=1}^{N} \mathbf{1}(y_i=c) \cdot \mathbf{1}(\hat{y}_i = c)
    \right)
\end{equation}

\begin{equation} \label{eq:baus}
    \text{BAUS}(y, \hat{y}) = \frac{1}{|I_U|} \sum_{c \in I_U}
    \left(
        \frac{1}{n_c} \sum_{i=1}^{N} \mathbf{1}(y_i=c) \cdot \mathbf{1}(\hat{y}_i = y_{\text{unknown}})
    \right)
\end{equation}

\begin{equation} \label{eq:final_score}
    \text{Score}(y, \hat{y}) = \sqrt{
        \text{BAKS}(y, \hat{y}) \cdot \text{BAUS}(y, \hat{y})
    }
\end{equation}

Our final score is then the geometric mean of the two measures.
Like the F1-score encourages models to balance precision and recall, the AnimalCLEF metric encourages models to be able to identify known and unknown individuals.

\subsection{Dataset Split for Open Set Classification}

The training images are stratified by individual into train, validation, and test sets, ensuring a robust and comprehensive dataset split.
The training set is used to fit a classification model, the validation set to observe progress across hyperparameter searches, and a test set for objective evaluation.
We organize the split in such a way that we can optimize a machine-learning algorithm to fit both the BAKS and BAUS objectives e.g. being able to be accurate about identifications of existing individuals and new individuals.

\begin{table}[h!]
\centering
\caption{
Dataset Split Summary. 
The train split is used to train the re-identification model to distinguish between known individuals. 
The validation split is for hyperparameter tuning and model selection. 
The test split is for final, unbiased performance evaluation.
In both validation and test splits, BAKS is calculated on the known individuals and BAUS on the unknown individuals.
}
\label{tab:dataset_split_revised}
\begin{tabular}{lcccc}
\hline
\textbf{Split} & \textbf{\shortstack{Num \\ Individuals}} & \textbf{\shortstack{Num \\ Images}} & \textbf{\shortstack{Known \\ Individuals}} & \textbf{\shortstack{Unknown \\ Individuals}} \\
\hline
Train      & 404 & 3392 & 404 & 0   \\
Validation & 458 & 2575 & 404 & 54  \\
Test       & 620 & 6568 & 404 & 216 \\
\hline
\end{tabular}
\end{table}

To predict the labels of existing individuals, we must ensure that there are known individuals shared between the training, validation, and test sets.
To predict unknown individuals, we select a set of individuals that are excluded from the training set but are known in the validation and test sets.
We describe the statistics of the train-validation-test split in table~\ref{tab:dataset_split_revised}.
We use 60\% of the training individuals for training, 20\% for validation, and the remaining 20\% for testing. 
If an individual has only a single image, it belongs to the training dataset by default.

\subsection{Transfer Learning via ViT Embedding Extraction}

We hypothesize that pre-trained self-supervised vision transformer models learn an adequate feature space for distinguishing individuals.
The images are projected onto a lower-dimensional manifold that roughly maps semantic distances found in the original space. 
The new points are called embeddings and are vectors of numbers that capture the lower dimensional latent space.
Embeddings capture semantic similarities between images through inductive biases of the model and the distribution of the training dataset.
Vision transformers learn to represent an image through a sequence of tokens derived from patches of the original image in addition to a special token called the classification (CLS) token.
We capture and transfer the underlying knowledge by extracting embeddings from the model by extracting the CLS token.

\begin{table}[h!]
\centering
\caption{Comparison of DINOv2 and MegaDescriptor Foundation Models}
\label{tab:model_comparison_vlines}
\begin{tabular}{|l|p{5.5cm}|p{5.5cm}|}
\hline
\textbf{Feature} & \textbf{DINOv2 (Base)} & \textbf{MegaDescriptor (Large)} \\
\hline
\textbf{Underlying Architecture} & Vision Transformer (ViT-B) & Swin Transformer (Swin-L) \\
\hline
\textbf{Parameter Size} & $\sim$87 Million & $\sim$229 Million \\
\hline
\textbf{Output Dimension} & 768 & 1024 \\
\hline
\textbf{Training Dataset} & LVD-142M: A large-scale, general-purpose dataset of 142 million images. & A collection of 29+ public animal re-identification datasets (e.g., WildlifeReID-10k). \\
\hline
\textbf{Specialization} & General-purpose vision foundation model. & Foundation model trained for wildlife re-identification. \\
\hline
\end{tabular}
\end{table}

\begin{figure}[h!]
    \centering
    
    \begin{subfigure}{0.47\textwidth}
        \centering
        \includegraphics[width=\linewidth]{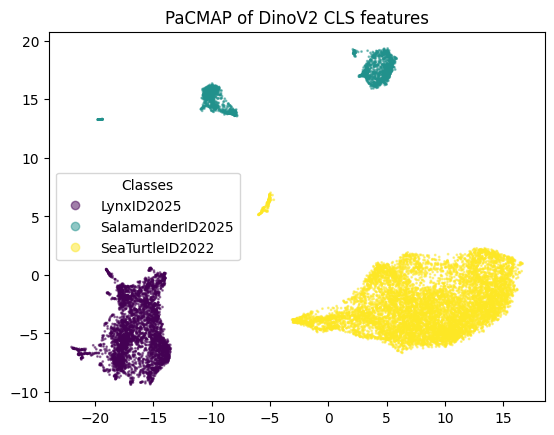}
        \caption{\footnotesize DINO CLS Embeddings with PaCMAP}
        \label{fig:dino_pacmap}
    \end{subfigure}
    \hfill 
    \begin{subfigure}{0.47\textwidth}
        \centering
        \includegraphics[width=\linewidth]{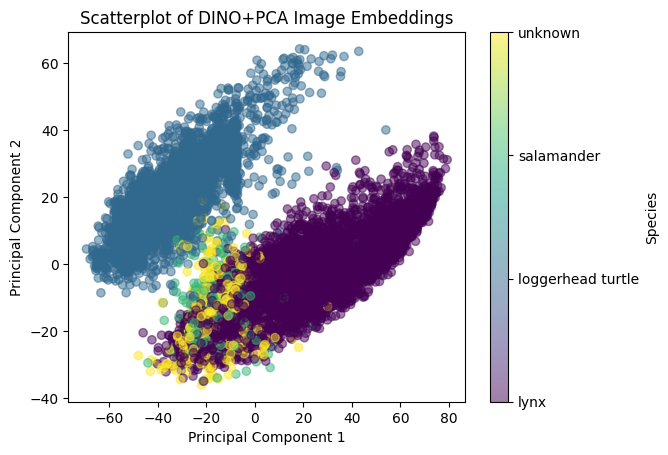}
        \caption{\footnotesize DINO CLS Embeddings with PCA}
        \label{fig:dino_pca}
    \end{subfigure}
    
    \caption{Comparison of dimensionality reduction techniques (PaCMAP vs. PCA) on the DINO CLS token embeddings from the same dataset.}
    \label{fig:reduction_comparison}
\end{figure}

We can demonstrate a degree of visual separation by projecting the embeddings onto a 2D manifold, which can be visualized as a scatter plot.
In figure~\ref{fig:reduction_comparison}, we embed the entire dataset using a pre-trained DINOv2 model from HuggingFace.
We then use principle component analysis (PCA) and pairwise controlled manifold approximation (PaCMAP \cite{pacmap2021}) to visualize how the points cluster in space.
PCA works by normalizing the vectors into zero-mean and unit-variance matrix and then finding the rotation of the matrix that minimizes the projection in a lower-rank space.
The first two dimensions correspond to the principal axes of rotation, as determined through an eigen-decomposition.
We find that this projection can clearly separate lynxes from sea turtles, with some overlap between lynxes and salamanders.
We compare this approach with PaCMAP, which takes into account both local and global geometry by constructing a graph sampled from the original data.
This embedding better captures nuances of the original space and separates images in the training dataset into lynxes, sea turtles, and salamanders.

\subsection{Nearest Neighbor Classification}

The nearest neighbor classification takes the embeddings and determines which individual is closest to the point.
Images in the training dataset are used as prototypes for making the classification.
We use Faiss \cite{douze2024faiss} to index all of the training image embeddings for queries.
We use the L2 distance between points to rank a query vector to all of the training vectors and return the top K points.
We look at the nearest point and determine whether this is a new individual by applying a global threshold to the points.
If the nearest point is too far, then this is a new point.
Otherwise, we return the mode of the top K identities.

We choose the threshold through a procedure that optimizes the competition metric.
The threshold is selected by searching 100 linearly spaced points within a range of 3 Median Absolute Deviations (MAD) from the median distance between each image and its nearest neighbor of a different species. 
The threshold that maximizes the competition score (geometric mean of BAKS and BAUS) on the validation set is chosen.
These statistics are robust indicators of the dataset and are less influenced by outliers.
The MAD is also a value that is independent of the domain of the thresholds, such that we can describe distances in their modified z-score in the distribution of distances. 

\subsection{Triplet Learning}

We also applied the triplet learning paradigm in order to learn a better representation of the data.
The objective of the triplet loss is to ensure that same-identity pairs are closer than those of different identities using an anchor-positive-negative triplet.

$$L = \max(d(x_a, x_p) - d(x_a, x_n) + \alpha, 0)$$

In this formulation, $x_a$ represents the embedding of an anchor image, $x_p$ is the embedding of a positive image from the same individual, and $x_n$ is the embedding of a negative image from a different individual.
The function $d$ calculates the L2 distance between two embeddings, and the hyperparameter $\alpha$ represents the margin that enforces separation between the pairs.
The loss is minimized only when the distance between the anchor and positive pair is smaller than the distance between the anchor and negative pair by at least the margin $\alpha$.
For our experiments, we followed a standard approach and utilized a unit margin where $\alpha=1$.

\begin{figure}[!htbp]
    \centering
    \includegraphics[width=1.0\linewidth]{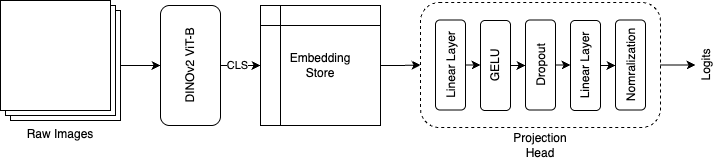}
    \caption{Our triplet learning pipeline. Raw images are pre-processed by a frozen DINOv2 ViT-B model, from which the 768-dimensional CLS embeddings are extracted and stored. From this store, the CLS embeddings are passed to the projection head, which is made up of a linear layer, GELU activation, dropout, a second linear layer, and L2 normalization.}
    \label{fig-triplet}
\end{figure}

We pre-compute embeddings with DINOv2 and MegaDescriptor-L-384 derived from the CLS token in the ViT.
These CLS embeddings were then downsampled by a projection head consisting of two linear layers with GELU activation and dropout sandwiched between them, followed by L2 normalization after the second linear layer. 
The first linear layer is equal to the size of the original embedding space, and the second layer is set to a value of 256.
The model must be parameterized in such a way that it can capture the relationship between triplets, given their locations in the original manifold, with the ability to generalize to new examples.
This pipeline is depicted in Figure \ref{fig-triplet}. 

The projection head was trained with a batch size of 200 on the CLS embeddings for the images in the database, split over 100 epochs in total. 
We experimented with using standard triplet loss \cite{facenet} as well as a modified triplet loss using Matryoshka Representation Learning \cite{mrl}, both with unit margins. 
We also experimented with different online triplet mining techniques, specifically random selection and semi-hard negative selection \cite{facenet}. 
The Adam optimizer was used with a learning rate of $5\times10^{-4}$; a linear scheduler was employed for warmup, followed by cosine annealing after the 10th epoch.

\section{Results}

We report our final private leaderboard score on Kaggle.
Our best model, which utilized MegaDescriptor with triplet loss, achieved a score of 0.39 and ranked 103 out of 174.
This is 0.09 points above the MegaDescriptor baseline provided by the competition organizers and 0.05 points below the WildFusion baseline.  

\begin{table}[h!]
\centering
\caption{
    Leaderboard rankings and scores.
    The rank is given by the private/public rankings, with the public rankings being out of 174 teams.
    The score is given by the competition metric.
    }
\label{tab:leaderboard}
\begin{tabular}{llcc}
\hline
\textbf{Rank} & \textbf{Name} & \textbf{Public} & \textbf{Private} \\
\hline
1/1 & DataBoom & 0.72781 & 0.71388 \\
2/2 & webmaking & 0.65832 & 0.67420 \\
3/3 & hzhzh & 0.63448 & 0.65753 \\
98/88 & WildFusion - MegaD. + ALIKED & 0.36555 & 0.44362 \\
100/103 & DS@GT LifeCLEF & 0.35583 & 0.39082 \\
126/128 & MegaDescriptor-L-384 & 0.30002 & 0.30898 \\
\hline
\end{tabular}
\end{table}

\begin{table}[h!]
\centering
\caption{Model performance of models submitted to the competition.}
\label{tab:model_scores_summary}
\begin{tabular}{lccc}
\hline
\textbf{Name} & \textbf{Public Score} & \textbf{Private Score} & \textbf{Submission Name} \\
\hline
dino baseline & 0.24371 & 0.18856 & 20250313-baseline.csv \\
megadescriptor baseline & 0.28528 & 0.25967 & 20250422-baseline.csv \\
dino linear & 0.27116 & 0.27990 & dino-semihard-epoch80-prediction.csv \\
dino nonlinear & 0.18515 & 0.21752 & nonlinear-v3-epoch20.csv \\
megadescriptor nonlinear & 0.35583 & 0.39082 & megadescriptor-nonlinear-128-epoch100.csv \\
\hline
\end{tabular}
\end{table}

We train several models with our described methodology in table~\ref{tab:model_scores_summary}.
Our baseline models are the result of embedding the data into either DINOv2 base or MegaDescriptor-L-384 and then applying our K-NN classification procedure with thresholding.
We train another model with the triplet learning embedding head, with a linear projection from the original embedding space down to a dimension of 256.
Finally, we compare a non-linear projection learned by the final methodology.

\begin{figure}[h!]
    \centering
    \includegraphics[width=\linewidth]{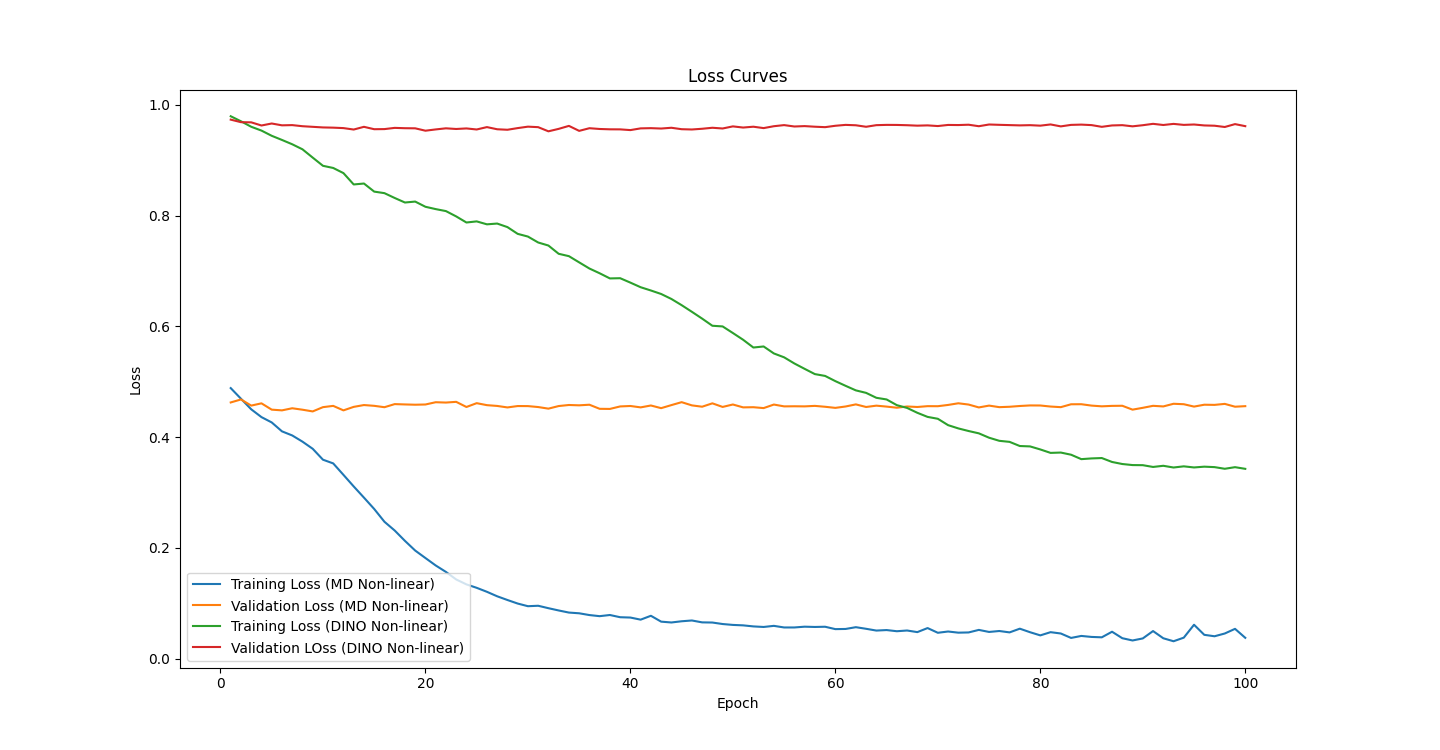}
    \caption{
        The training losses for the non-linear triplet mined embeddings.
        We observe that the MegaDescriptor head achieves significantly lower training and validation losses compared to Dino.
    }
    \label{fig:all-losses}
\end{figure}

\begin{figure}[h!]
    \centering
    \includegraphics[width=1\linewidth]{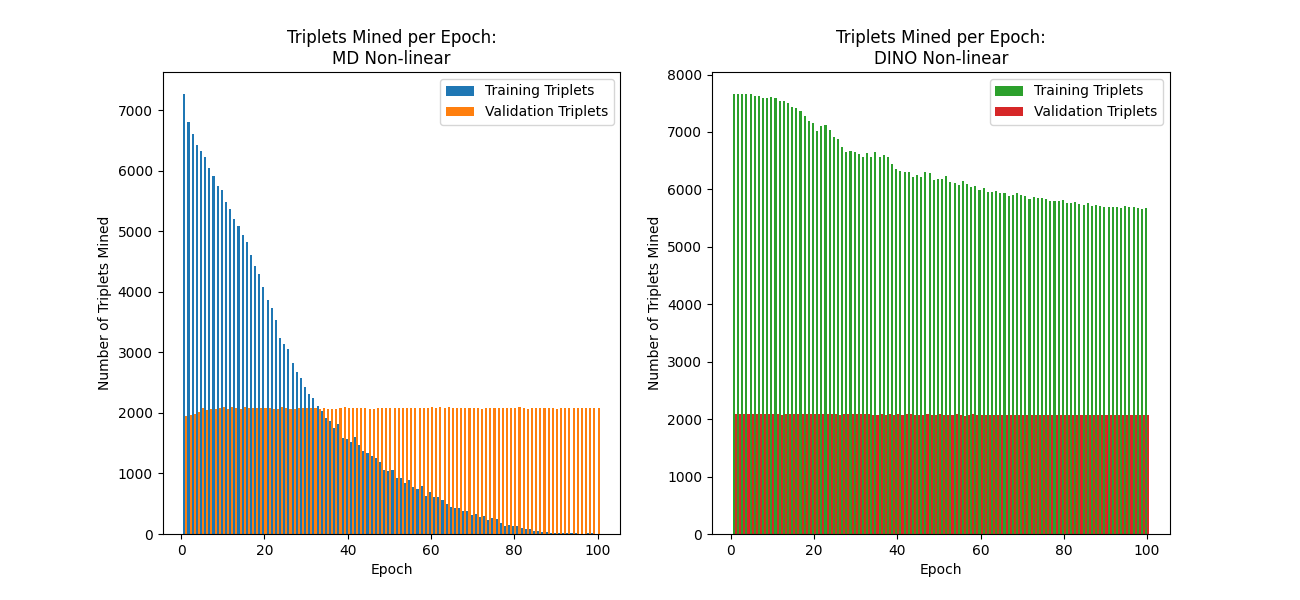}
    \caption{
    The number of triplets mined at epoch using a semi-hard negative mining routine.
    We note that MegaDescriptor can reduce the number of triplets that satisfy the semi-hard margin constraints over the 100-epoch training loop to a much larger degree than Dino.
    }
    \label{fig:all-triplets}
\end{figure}

In addition to the final results, we also report some of the training dynamics of the triplet learning procedure to illustrate the differences between DINOv2 and MegaDescriptor.
In figure~\ref{fig:all-losses}, we see that the triplet loss objective is lower across all epochs in MegaDescriptor over Dino.
Both models decrease in loss over the 100 epoch during training, meaning that fewer triplets are violating the margin constraint during training, as observed by the valid triplets found in figure~\ref{fig:all-triplets}.
However, we note that while the training loss continues to decrease, the validation loss converges quickly. 

\begin{figure}[h!]
    \centering
    
    \begin{subfigure}{0.48\textwidth}
        \centering
        \includegraphics[width=\linewidth]{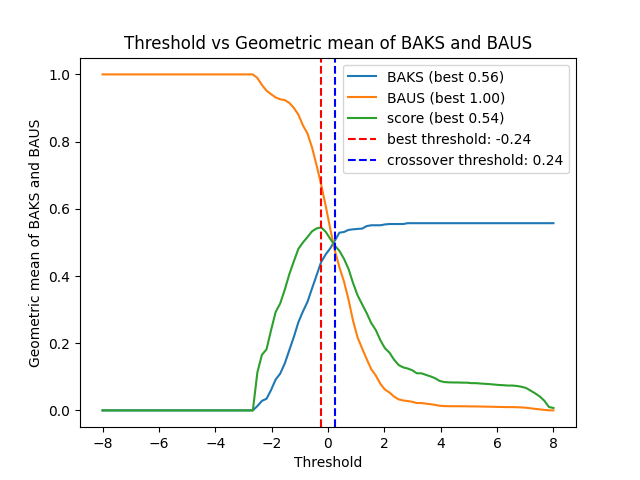}
        \caption{DINO Non-Linear Validation Curve}
        \label{fig:dino_threshold_curve}
    \end{subfigure}
    \hfill 
    \begin{subfigure}{0.48\textwidth}
        \centering
        \includegraphics[width=\linewidth]{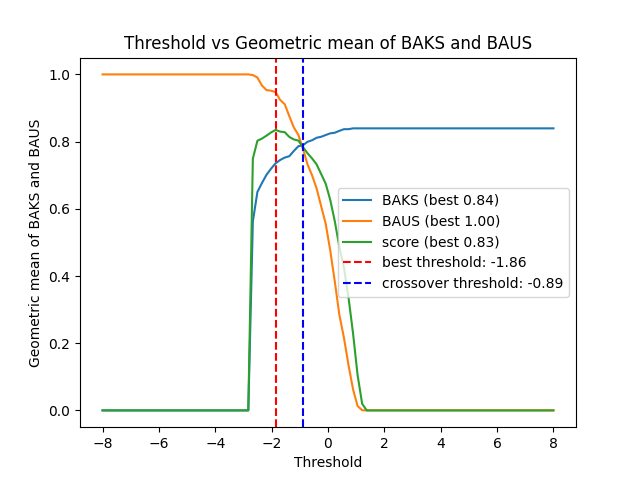}
        \caption{MegaDescriptor Non-Linear Validation Curve}
        \label{fig:megadescriptor_threshold_curve}
    \end{subfigure}
    
    \caption{Validation curves used to determine the optimal classification threshold for each model.}
    \label{fig:threshold_curves}
\end{figure}

Finally, we report the hyperparameter tuning of the k-NN classification threshold in figure~\ref{fig:threshold_curves}.
For the triplet training epoch with the best validation loss, we run our tuning procedure to find the best threshold. 

\section{Discussion}

\begin{figure}[h!]
    \centering
    \begin{subfigure}{0.48\textwidth}
        \centering
        \includegraphics[width=\linewidth]{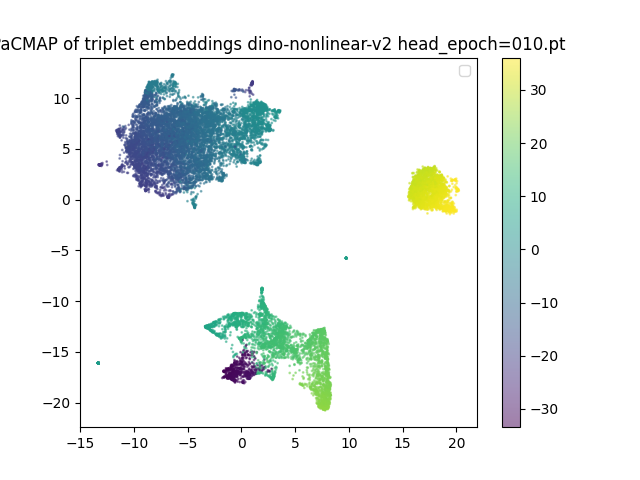}
        \caption{DINO Non-Linear Embedding, Epoch 10}
        \label{fig:dino_embedding_e10}
    \end{subfigure}
    \hfill 
    \begin{subfigure}{0.48\textwidth}
        \centering
        \includegraphics[width=\linewidth]{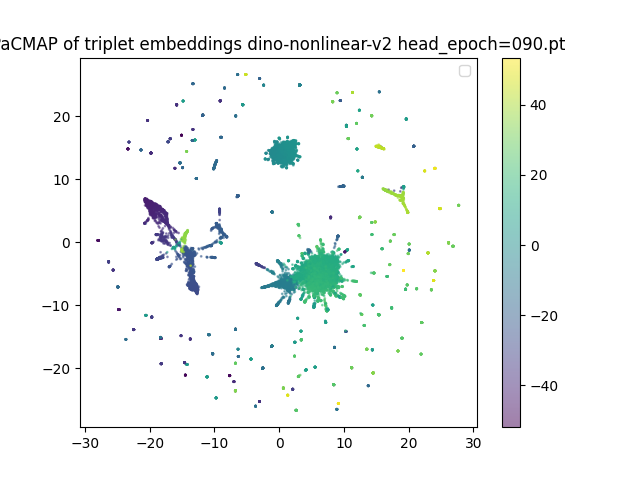}
        \caption{DINO Non-Linear Embedding, Epoch 90}
        \label{fig:dino_embedding_e90}
    \end{subfigure}
    \begin{subfigure}{0.48\textwidth}
        \centering
        \includegraphics[width=\linewidth]{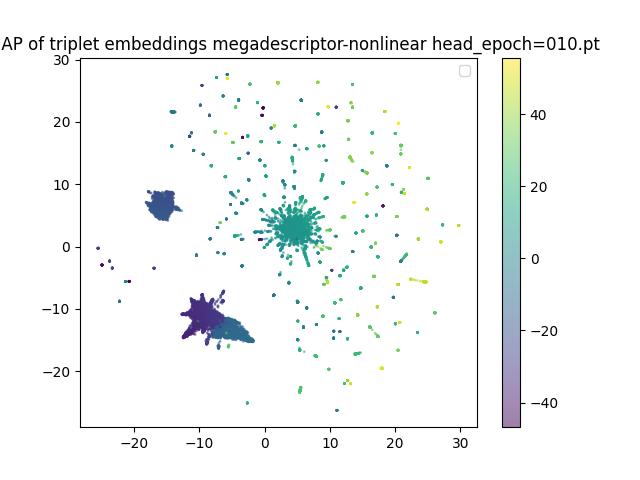}
        \caption{MegaDescriptor N.L. Embedding, Epoch 10}
        \label{fig:megadescriptor_embedding_e10}
    \end{subfigure}
    \hfill 
    \begin{subfigure}{0.48\textwidth}
        \centering
        \includegraphics[width=\linewidth]{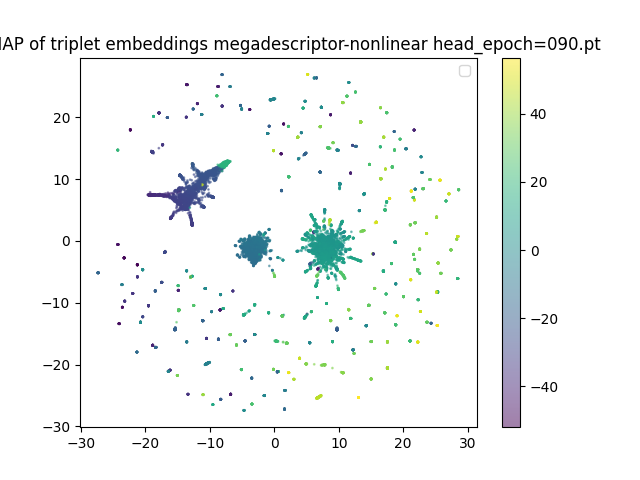}
        \caption{MegaDescriptor N.L. Embedding, Epoch 90}
        \label{fig:megadescriptor_embedding_e90}
    \end{subfigure}
    
    \caption{Visualization of triplet embedding spaces for DINO and MegaDescriptor models at early (Epoch 10) and late (Epoch 90) stages of training.}
    \label{fig:embedding_space_grid}
\end{figure}

Through our experiments, we found that domain-specific training is crucial for achieving good performance on re-identification tasks. 
We see this in our baseline k-NN behavior against the DINOv2 and MegaDescriptor embeddings.
We observe a 0.07-point difference on the private leaderboard between these two models, with the MegaDescriptor model performing better.

Our procedure is 0.04 points lower than the competition baseline despite using the same model.
This may be caused by improper resizing before calling the model or by selecting an inappropriate metric.
The starter notebook uses cosine distance with a fixed threshold of 0.6, while we use Euclidean distance with a threshold chosen by a hyperparameter search influenced by our dataset split.
The cosine distance is the more appropriate measure, and requires using an inner-product index with unit-normalized vectors.
It is also possible that the implicit assumptions in the dataset split had a substantial impact on the distribution of outputs. 
We use 60\% of the dataset as "known" in our experiments, but it is possible that increasing the set of known images would lead to a different optimal hyperparameter score.
While we try to have an offline approximation of the private leaderboard for development, it has proven challenging to find a suitable proxy for local development. 

During development, we also found that setting up the triplet loss was particularly tricky.
Although we employed a train-validation-test split for our thresholding scheme, we opted for a distinct train-validation split for the triplet learning pipeline.
While we could continue to learn and reduce the triplet loss, it also indicated that we were overfitting the geometry of the training dataset.
When we accounted for our split, we had a better chance of determining which parameterization of the triplet layer was most effective for us. 

After refining our dataset split using our triplet pipeline, we found that reshaping DINOv2 was significantly harder than MegaDescriptor.
DINOv2 is a general-purpose feature extractor, and while it performs well on generalized tasks, it is not optimized for fine-grained classification in this context.
We did not find a mining strategy or parameterization of the embedding head that would lower the triplet loss on the validation set.
Since the training and validation sets are non-overlapping, we can only indirectly influence the triplet scores of the validation set by reshaping the manifold through triplet mining on the training set.
Individuals are not clustered as tightly on the DINOv2 manifold, and it becomes difficult to move in a direction that is isomorphic between training and validation sets.
The MegaDescriptor triplet learning works comparably well, increasing the model's performance on the task by 0.13, compared to a 0.03 improvement on DINOv2 triplet learning.
This could be because the MegaDescriptor model employs a metric objective that combines ArcFace and Triplet Loss, resulting in a minor domain shift overall.

Finally, we observe differences in the triplet learning in figure~\ref{fig:embedding_space_grid}.
We use PaCMAP, a graph-theoretic embedding method that takes into account both local and global geometry in shape.
The large cluster thus signifies a cloud of points that are challenging to disambiguate.
We see that over the process of triplet learning, the DINO model learns to disambiguate clusters of individuals.
In the MegaDescriptor model, there is already a large number of clusters.
Note that the number of clusters is visually larger than in the DINO model, and this roughly correlates with performance in the final task.

\section{Future Work}

Reflecting on the poor performance achieved using DINO, it would likely have performed better if we had supplemented training with the WildlifeReID-10K dataset \cite {wildlifereid10k}.
Previous experience with similarly designed pipelines has made us aware of the data-hungry nature of the triplet learning paradigm. 
Due to the small size of the provided dataset and the limitations imposed on our triplet mining implementation, the number of unique triplets used during training was likely insufficient. Using the WildlifeReID-10K dataset in conjunction with the provided data would likely alleviate these issues.

\begin{table}[h!]
\centering
\caption{Comparison of Model Parameters for ViT Backbones}
\label{tab:model_params}
\begin{tabular}{l r}
\hline
\textbf{Model Name} & \textbf{Parameters (Millions)} \\
\hline
\texttt{facebook/dinov2-small} & 21 \\
\texttt{facebook/dinov2-base} & 86 \\
\texttt{facebook/dinov2-large} & 304 \\
\texttt{facebook/dinov2-giant} & 1,135 \\
\texttt{BVRA/MegaDescriptor-T-224} & 28.3 \\
\texttt{BVRA/MegaDescriptor-S-224} & 49.6 \\
\texttt{BVRA/MegaDescriptor-B-224} & 109.1 \\
\texttt{BVRA/MegaDescriptor-L-224} & 228.6 \\
\texttt{BVRA/MegaDescriptor-L-384} & 228.8 \\
\texttt{BVRA/MegaDescriptor-T-CNN-288} & 12.2 \\
\hline
\end{tabular}
\end{table}

Additionally, we would like to experiment with a larger number of backbones to ensure that results are comparable. 
We enumerate a list of models in Table~\ref{tab:model_params} that would provide a concrete starting point for future experiments.

\section{Conclusions}

We develop a transfer learning solution for the AnimalCLEF 2025 competition, leveraging inherent visual knowledge encoded in vision transformers.
We define the nearest neighbor classifier that can tackle the open-set nature of the competition through a rigorously defined thresholding procedure.
While our solution ranks higher than the baseline MegaDescriptor solution, there are limitations to our methods that should be addressed by augmenting them with a larger individual dataset and more careful hyperparameter tuning.
Code for this paper can be found at \url{https://github.com/dsgt-arc/animalclef-2025}.

\section*{Acknowledgements}

We thank the Data Science at Georgia Tech (DS@GT) CLEF competition group for their support.
This research was supported in part through research cyberinfrastructure resources and services provided by the Partnership for an Advanced Computing Environment (PACE) at the Georgia Institute of Technology, Atlanta, Georgia, USA \cite{PACE}. 

\section*{Declaration on Generative AI}

 During the preparation of this work, the authors used Gemini Pro and Grammarly in order to: Abstract drafting, formatting assistance, grammar and spelling check. After using these tools/services, the authors reviewed and edited the content as needed and takes full responsibility for the publication’s content. 
\bibliography{main}
\end{document}